# Multi-Agent Imitation Learning for Driving Simulation

Raunak P. Bhattacharyya,[*,1] Derek J. Phillips,[*,1] Blake Wulfe,[1] Jeremy Morton,[1]
Alex Kuefler,[2] and Mykel J. Kochenderfer[1]

*Abstract*—Simulation is an appealing option for validating the safety of autonomous vehicles. Generative Adversarial Imitation Learning (GAIL) has recently been shown to learn representative human driver models. These human driver models were learned through training in single-agent environments, but they have difficulty in generalizing to multi-agent driving scenarios. We argue these difficulties arise because observations at training and test time are sampled from different distributions. This difference makes such models unsuitable for the simulation of driving scenes, where multiple agents must interact realistically over long time horizons. We extend GAIL to address these shortcomings through a parameter-sharing approach grounded in curriculum learning. Compared with single-agent GAIL policies, policies generated by our PS-GAIL method prove superior at interacting stably in a multi-agent setting and capturing the emergent behavior of human drivers.

## I. INTRODUCTION

Validating the safety of autonomous vehicles represents an unsolved yet crucial challenge for regulators and manufacturers alike. Autonomous driving systems are typically evaluated on real-world drive tests, which are expensive, time-consuming, and potentially dangerous. Furthermore, it is likely infeasible to build a statistically significant case for the safety of a system solely through real-world testing [1], [2]. Validation through simulation provides a promising alternative to real-world testing, with the ability to evaluate vehicle performance in large numbers of scenes quickly, safely, and economically.

Simulations must accurately reflect real-world driving to be useful, and therefore require realistic models of human drivers to govern the behavior of non-autonomous vehicles that occupy the roadway.

Imitation learning (IL) represents a promising avenue for learning models of driver behavior from real-world data. Behavioral cloning, a variant of imitation learning that relies on a supervised learning procedure, is easy to implement but tends to perform poorly in practice due to a problem known as covariate shift [3]. Inverse Reinforcement Learning (IRL) approaches formulate the imitation learning task as a Markov Decision Process (MDP) with a stationary dynamics model, which can be solved approximately using batch policy optimization [4]. IRL addresses the covariate shift problem by learning a cost function and allowing the agent to interact with the environment while training. As a result, the agent encounters similar states during training and testing, which in practice enables the agent to better learn the consequences of—and correct for—its mistakes.

Imitation learning approaches have been shown to successfully learn human driving policies for individual vehicles in car-following and highway-driving contexts [5], [6]. However, when evaluated in a multi-agent setting, the policies learned through single-agent imitation learning fail to exhibit realistic behavior, rendering them inadequate for use in simulation. As we argue later, this deterioration in performance occurs because transitioning from single-agent to multi-agent settings effectively reintroduces the covariate shift problem.

We derive an algorithm that addresses the deficiencies of single-agent imitation learning for learning human driver models. We extend Generative Adversarial Imitation Learning (GAIL) [7] and Parameter Sharing Trust Region Policy Optimization (PS-TRPO) [8] to enable imitation learning in the multi-agent context, yielding a new algorithm called PS-GAIL. PS-GAIL generates policies capable of controlling multiple vehicles simultaneously, enabling the simulation of complex roadway scenes.

The effectiveness of the proposed PS-GAIL method is demonstrated by comparing the performance of its learned policies to single-agent policies learned through GAIL. PS-GAIL policies are shown to generate driving trajectories that better match those of human drivers. Furthermore, vehicles driven by PS-GAIL policies are shown to interact with each other in a more stable manner over long horizons, and they are less prone to the collisions and off-road events that often arise during interactions between single-agent GAIL policies.

## II. BACKGROUND

### A. Markov Decision Processes

An infinite horizon MDP is defined by the tuple $(\mathcal{S}, \mathcal{A}, \mathcal{T}, \mathcal{R}, \gamma)$, where $\mathcal{S}$ is the state space, $\mathcal{A}$ is the action space, $\mathcal{T}$ is the transition model, $\mathcal{R}$ is the reward function, and $\gamma$ is the discount factor. The reward function $\mathcal{R}$ provides the rewards received while interacting in the environment, where $\mathcal{R}(s, a, s')$ denotes the reward for transitioning from $s$ to $s'$ when action $a$ is taken. The transition model $\mathcal{T}(s' \mid s, a)$ gives the probability over next states given action $a$ taken in state $s$. The discount factor governs how much future rewards are valued relative to immediate rewards.

A stochastic policy, $\pi : \mathcal{S} \to P(\mathcal{A})$, maps each state to the probability of taking each action. The sum of discounted rewards, or return, from state $s_t$ is defined as $g_t = \sum_{k=0}^{\infty} \gamma^k r_{t+k+1}$, where $t$ is a time index, and $r_t$ is the corresponding reward. The objective in an MDP is to find a policy that maximizes the expected return, or value, of each state $V_\pi(s) = \mathbb{E}_\pi[g_t \mid s_t = s]$.

[*]Indicates equal contribution.
[1] Stanford Intelligent Systems Laboratory, Stanford University, Stanford, CA 94305, USA, {raunakbh, djp42, wulfebw, jmorton2, mykel}@stanford.edu
[2] Osaro Inc., 118 2nd Street, Suite 200, San Francisco, CA 94105, USA, alex@osaro.com

When the agent only receives partial information about the state at each timestep, the problem can be formally represented as a partially observable Markov decision process (POMDP). A POMDP adds an observation space $\mathcal{O}$ and observation model $\mathcal{Z} : \mathcal{S} \to P(\mathcal{O})$, which provides the probability of the agent observing observation $o$ in state $s$ [9], [10].

### B. Imitation Learning

The goal in imitation learning (IL) is to learn a policy $\pi$ that imitates an expert policy $\pi_E$ given demonstrations from that expert [11], [12]. A demonstration is defined as a sequence of state-action pairs that result from a policy interacting with the environment: $d = \{s_1, a_1, s_2, a_2, \ldots\}$. In this setting, the reward function is typically assumed to be unknown.

One particularly useful definition in imitation learning is that of the state-occupancy distribution:

$$\rho_\pi(s) = (1 - \gamma) \sum_{t=0}^{\infty} \gamma^t p(s_t = s \mid \pi), \quad (1)$$

which gives the average discounted probability of the agent being in state $s$. The supervised learning approach to imitation learning, behavioral cloning (BC), learns a policy by minimizing some loss function $\ell$ over the set of demonstrations with respect to the policy [12]:

$$\pi_{sup} = \underset{\pi}{\arg\min} \underset{s \sim \rho_{\pi_E}}{\mathbb{E}} [\ell(\pi, s)], \quad (2)$$

where $\ell$ is typically the cross-entropy loss when using discrete actions and the negative log likelihood of a multivariate Gaussian distribution when using continuous actions.

During training, behavioral cloning samples states from the state-occupancy distribution of the expert, $\rho_{\pi_E}$. However, when interacting with the environment, the policy samples states from the state-occupancy distribution of the learned policy, $\rho_{\pi_{sup}}$. This change in distribution between training and test time is referred to as covariate shift [3], and results in the agent making increasingly large errors from which it cannot recover.

Allowing the agent to interact with the environment during training time addresses the underlying cause of covariate shift, but this interaction requires a reward function since the agent may encounter states not contained in the training data. While different methods exist for addressing this problem, we focus on Generative Adversarial Imitation Learning (GAIL) due to its scalability, low sample-complexity as measured in expert demonstrations [7], and previous success in learning human driver models [6].

GAIL formulates imitation learning as the problem of matching the state-action occupancy distribution of the expert policy. Ho et al. [7] show that generative adversarial networks can be used to approximately accomplish this task. A discriminator $D_\psi$, parametrized by $\psi$ learns to distinguish expert from non-expert behavior, while a policy $\pi_\theta$, parameterized by $\theta$ attempts to emulate that behavior. In the case where $D_\psi$ represents the probability the state-action pair came from $\pi_E$, the GAIL objective is given by [13]:

$$\min_\theta \max_\psi \mathbb{E}_{\pi_E} \log D_\psi(s, a) + \\ \mathbb{E}_{\pi_\theta} \log(1 - D_\psi(s, a)). \quad (3)$$

Minimizing different divergences between state-action occupancy distributions yields different objectives. We use the Wasserstein distance [14] because it mitigates the vanishing gradient problem observed when minimizing Jensen-Shannon divergence [14] and because it empirically yields better policies. The objective becomes:

$$\min_\theta \max_\psi \mathbb{E}_{\pi_E}[D_\psi(s, a)] - \mathbb{E}_{\pi_\theta}[D_\psi(s, a)], \quad (4)$$

where the critic, $D$, learns to output a high score when encountering pairs from $\pi_E$, and a low score when encountering generated pairs. The output of the critic $D_\psi(s, a)$ can then be used as a surrogate reward function whose value grows larger as actions sampled from $\pi_\theta$ look similar to those chosen by experts.

After performing rollouts with a given set of policy parameters $\theta$, surrogate rewards $\tilde{r}(o, a; \psi)$ are calculated and Trust Region Policy Optimization (TRPO) [15] is used to perform a policy update. TRPO is used because it can better handle the high variance of the non-stationary reward from the critic. Although $\tilde{r}(o, a; \psi)$ may be quite different from the true reward function optimized by experts, it can be used to drive $\pi_\theta$ into regions of the state-action space similar to those explored by $\pi_E$.

GAIL effectively addresses the problem of covariate shift provided that the training and testing environments are identical; however, this is not the case when learning human driver models in a single-agent setting and deploying them in a multi-agent setting. During test time, the policy observes nearby vehicles acting differently than during training, and again makes small errors that compound over time. In this paper, we explicitly formulate learning human driver models as a multi-agent problem in order to address this discrepancy.

### C. Multi-agent reinforcement learning

The task of simultaneously controlling multiple vehicles operating on a single roadway can be viewed as a multi-agent control problem. Multi-agent control has been studied in extensive detail from the dynamical systems perspective in problems like coverage control [16], formation control [17], and consensus [18]. The dynamical systems approach allows for development of provable characteristics about the controller, but requires problem specific features and hand-engineered control laws. In contrast with such multi-agent control approaches, the field of multi-agent reinforcement learning offers a more general way to solve multi-agent problems [19], [20].

Centralized multi-agent RL traditionally requires learning a policy $\pi$ mapping from a joint observation space $\mathcal{O} \in \mathbb{R}^{O \times M}$ to action space $\mathcal{A} \in \mathbb{R}^{A \times M}$. Here $O$ is the dimensionality of a single observation, $M$ is the number of agents, and

$A$ is the dimensionality of a single action. Centralized approaches grow rapidly in computational complexity as the number of agents increases. Provided that the pool of agents is homogeneous (i.e., each policy must perform essentially the same mapping, $\mathcal{O} \to \mathcal{A}$), the joint distribution $\pi = p(a_0, \ldots, a_M \mid o_0, \ldots, o_M)$ can be factorized as $\prod_i p(a_i \mid o_i)$, which scales linearly with number of agents.

Gupta, Egorov, and Kochenderfer introduced an algorithm called Parameter Sharing Trust Region Policy Optimization (PS-TRPO), which is a policy gradient approach that combines parameter sharing and TRPO [8]. PS-TRPO was shown to produce decentralized parameter sharing neural network policies that exhibit emergent cooperative behavior without explicit communication between agents. PS-TRPO is highly sample-efficient because it reduces the number of parameters by a factor of $M$, and shares experience across all agents. Notably, PS-TRPO still allows agents to exhibit different behavior because each agent receives unique observations.

For a policy $\pi_\theta$ with parameters $\theta$, PS-TRPO performs an update to the policy parameters by approximately solving the constrained optimization problem:

$$\underset{\theta}{\text{maximize}} \quad \mathbb{E}_{o, a \sim \pi_{\theta_k}} \left[ \frac{\pi_\theta(a \mid o)}{\pi_{\theta_k}(a \mid o)} A_{\theta_k}(o, a) \right] \quad (5)$$
$$\text{subject to} \quad \mathbb{E}_o[D_{KL}(\pi_{\theta_k}(\cdot \mid o) \| \pi_\theta(\cdot \mid o))] \leq \Delta_{KL},$$

where $\pi_{\theta_k}$ is a rollout-sampling policy and $A_{\theta_k}(o, a)$ is an advantage function quantifying how much the value of an action $a$ taken in response to an observation $o$ differs from the baseline value estimated for $o$. $D_{KL}$ is the KL divergence between the two policy distributions, and $\Delta_{KL}$ is a step size parameter that controls the maximum change in policy per optimization step.

### III. APPROACH

We propose an extension to GAIL enabling the simultaneous control of multiple human driver models. The following subsections describe the multi-agent driving problem, and detail our approach for combating its associated challenges.

#### A. Problem Formulation

In line with recent work in multi-agent imitation learning [21], we formulate multi-agent driving as a Markov game [22] consisting of $M$ agents and an unknown reward function. We make three simplifying assumptions:

1) **Homogeneous agents**: agents have the same observation and action spaces:

$$\mathcal{O}_i = \mathcal{O}_j \text{ and } \mathcal{A}_i = \mathcal{A}_j \ \forall \text{ agents } i, j.$$

2) **Independent rewards**: the reward function is not shared; it depends only on the action of each agent and the state, and not on the actions of other agents or the next state. In particular, agents are not cooperative:

$$\mathcal{R}_i(s, a_1, \ldots, a_i, \ldots, a_k) = \mathcal{R}_i(s, a_i).$$

3) **Identical reward function**: the reward function is the same for all agents:

$$\mathcal{R}_i = \mathcal{R}_j \ \forall \text{ agents } i, j.$$

---

**Algorithm 1** PS-GAIL

**Input:** Expert trajectories $\tau_E \sim \pi_E$, Shared policy parameters $\Theta_0$, Discriminator parameters $\psi_0$, Trust region size $\Delta_{KL}$
**for** $k \leftarrow 0, 1, \ldots$ **do**
    Rollout trajectories for all agents $\vec{\tau} \sim \pi_{\theta_k}$
    Score $\vec{\tau}$ with critic, generating reward $\tilde{r}(s_t, a_t; \psi_k)$
    Batch trajectories obtained from all the agents
    Take a TRPO step to find $\pi_{\theta_{k+1}}$ maximizing Eq. (5)
    Update the critic parameters $\psi$ by maximizing Eq. (4)
**end for**

---

These assumptions are idealizations and do not hold for real-world driving scenes. For example, different vehicles may permit different accelerations, a driver may only want to change lanes if other drivers are not doing so, and individuals may value different driving qualities such as smoothness or proximity to other vehicles differently. Nevertheless, these assumptions often do apply approximately, and, as we later show, allow for learning of realistic driving policies.

#### B. Parameter Sharing GAIL

A naive approach to learning human driver policies would be to train a policy in an environment where it controls a single vehicle on the roadway and all remaining vehicles follow a predetermined trajectory. Unfortunately, this approach is often incapable of producing policies that can reliably control many vehicles on the same roadway. By introducing such a controller to other vehicles after training, we reintroduce covariate shift. As a result, small errors in the behavior of a single vehicle can destabilize neighboring vehicles, ultimately leading to the failure of many agents in the scene.

Our proposed approach, PS-GAIL, combines GAIL with PS-TRPO to generate policies capable of driving multiple vehicles, enabling more stable simulation of entire road scenes. Algorithm 1 describes the PS-GAIL approach. We initialize the shared parameters of the policy and select a step size parameter. At each iteration of the algorithm, the policy with shared parameters is used by each agent to generate trajectories. Rewards are then assigned to each state-action pair in these trajectories by the critic. Subsequently observed trajectories are used to perform a TRPO update for the policy, and an Adam update for the critic. PS-GAIL can be viewed as a special case of the algorithms presented by Song et al. [21]. In particular, in PS-GAIL all agents share the same policy and receive rewards from the same critic.

We represent the policy with a recurrent neural network due to the high-dimensional observation space, nonlinearity required in the mapping from observations to actions, and partial observability of the local driving scene. Partial observability arises from (i) sensor noise and occlusions and (ii) unobserved driver latent state in the form of behavioral traits and intended maneuvers.

Our training procedure must also account for non-stationary environment dynamics. In the multi-agent setting, the dynamics of the environment change along with the agent policies.

We mitigate this problem by introducing a curriculum, $\mathcal{C}$, which scales the difficulty of the multi-agent learning problem during training. Gupta et al. define a multi-agent curriculum as a multinomial distribution over the number of agents controlled by the policy each episode: $\mathcal{C} \sim \text{Multi}(M, \vec{p})$ [8]. The curriculum gradually shifts probability mass to larger numbers of agents. In practice, we use a simplified curriculum that increments the number of controlled agents by a fixed number every $K$ iterations during training.

## IV. IMPLEMENTATION

This section describes the simulator used to train the multi-agent policies. Additionally, it details the implementation and training of the policy and critic networks.

### A. Simulator

In order to learn the policy in an environment with human drivers, we use a simulator that allows for playing back real trajectories and simulating the movement of controlled vehicles given actions selected by a policy. This process proceeds as follows:

1) The initial scene state is sampled from a dataset of real driver trajectories. This state includes the position, orientation, and velocity of all vehicles in the scene. The trajectory data we use is from the Next-Generation Simulation (NGSIM) dataset. NGSIM contains highway driving trajectories for US Highway 101 [23] and Interstate 80 [24], and consists of 45 minutes of driving at 10 Hz for each roadway.
2) A subset of the vehicles in the scene are randomly selected to be controlled by the policy. For single-agent training only one vehicle is selected, whereas for multi-agent training $M$ vehicles are controlled by the policy.
3) For each vehicle, a set of features are extracted and passed to the policy as the observation. Table I describes the features provided to the policy.
4) The policy outputs longitudinal acceleration and turn-rate values as the vehicle action. These values are used to propagate the vehicle forward in time.
5) This process repeats for a horizon of 200 timesteps at 10 Hz, corresponding to 20 s of driving per episode.

### B. Policies

We use recurrent neural network (RNN) policies, in all cases consisting of 64 Gated Recurrent Units (GRUs). The observation is passed directly into the RNN without any initial reduction in dimensionality. We use recurrent policies in order to address the partial observability of the state caused by occluded vehicles. In the multi-agent setting, a single shared policy selects actions for all vehicles, following the parameter sharing approach previously described. Policy optimization is performed using an implementation of TRPO from rllab [25], with a step size of 0.1.

We use two training phases for all of the models. The first phase consists of 1000 iterations with a low discount of 0.95 and a small batch size of 10 000 observation-action pairs. The second phase fine-tunes the models, running for 200 iterations with a higher discount of 0.99 and larger batch size of 40 000. For the multi-agent model, we add 10 agents to the environment every 200 iterations of the first training phase. We use 100 agents in the fine-tune phase for the multi-agent GAIL models.

TABLE I: Observation features

| Feature | Description |
| --- | --- |
| LIDAR Range and Range Rate | 20 artificial LIDAR beams output in regular polar intervals, providing the relative position and velocity of intercepted objects. |
| Ego Vehicle | Lane-relative velocity, heading, offset. Vehicle length and width. Lane curvature, distance to left and right lane makers and road edges. |
| Temporal | Longitudinal and lateral acceleration, local and global turn and angular rate, timegap, and time-to-collision. |
| Indicators | Collision occurring, ego vehicle out-of-lane, and negative velocity. |
| Leading Vehicle | Relative distance, velocity, and absolute acceleration of vehicle in front of fore vehicle, if it exists. |

### C. Critic

The critic acts as the surrogate reward function in the environment. The observation-action pairs for each vehicle at each timestep are passed to the critic, which outputs a scalar value that is then used as the reward for that vehicle. The critic is implemented as a feed-forward neural network consisting of (128,128,64) ReLU units. We implemented the critic as a Wasserstein GAN with gradient penalty (WGAN-GP) with a gradient penalty of 2 [26]. Similarly to Li et al. [27], we used a replay memory for the critic in order to stabilize training, which contains samples from the three most recent epochs. For each training epoch of the policy, the critic is trained for 40 epochs using the Adam optimizer [28] with a learning rate of 0.0004, dropout probability of 0.2, and batch size of 2000. Half of each batch consists of NGSIM data, with the remaining half comprised of data from policy rollouts. Finally, the reward values output from the critic are adaptively normalized to have zero mean and unit variance prior to being passed to TRPO.

Because the critic is a feed-forward network and operates on the observations of each vehicle, it is not able to account for the partial observability of the scene. A recurrent critic could address this issue, and has been used in other partially observable settings [29]. We nevertheless use a feed-forward critic because when using a recurrent critic one must decide when to output the reward–at each timestep or only at the end of the sequence. In the former case, the critic outputs large negative values at the beginning of a sequence when an agent closely resembles a real driver (i.e., it overfits), which makes learning difficult. In the latter case, the agent is faced with an extremely challenging credit assignment problem, receiving feedback on 200 timesteps of actions only at the terminal state. Li *et al.* provide a solution for the every-step reward problem, but it entails a highly computationally expensive Monte Carlo sampling procedure [29]. In practice, the feed-forward critic was sufficient to learn good driving policies, so we leave recurrent critics to future work.

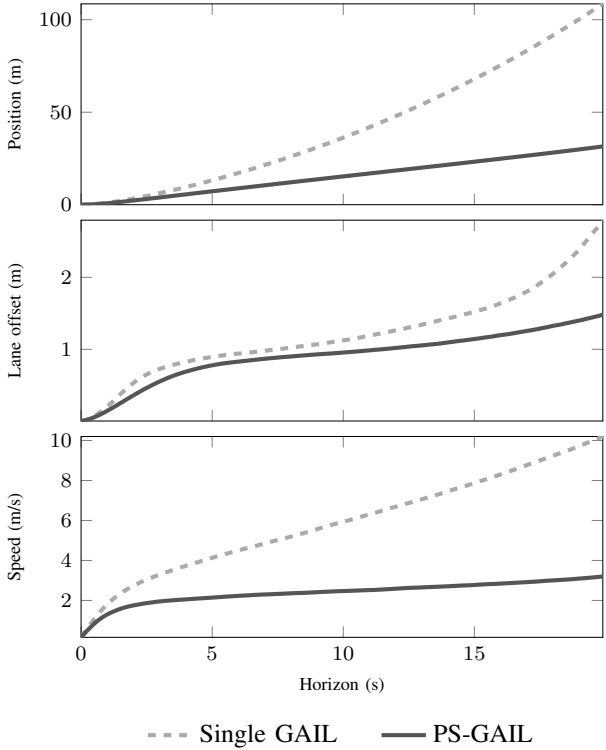

Fig. 1: Root weighted square error vs. prediction horizon for single and multi-agent models. The policy trained in a multi-agent setting more closely resembles the human driver, particularly as the horizon increases.

## V. EXPERIMENTS

We use GAIL and multi-agent GAIL to learn policies for two-dimensional highway driving and compare their performance. For all of the results presented, we train three of each model and average the results of 10 000 policy rollouts in a 100-agent environment. We present the average results over these rollouts.

### A. Evaluation Metrics

The trajectories generated by the policies are evaluated against human driving data using Root Mean Square Error (RMSE) and emergent behavior. For each of $m$ trajectories, we sample a single rollout and compute the RMSE of each predicted variable $v$:

$$\text{RMSE} = \sqrt{\frac{1}{m}\sum_{i=1}^{m}\left(v_t^{(i)} - \hat{v}_t^{(i)}\right)^2}, \quad (6)$$

where $v_t^{(i)}$ is the true value in the $i$th trajectory at time horizon $t$ and $\hat{v}_t^{(i)}$ is the simulated value for the $i$th trajectory at time horizon $t$. We extract the RMSE in predictions of global position, lane offset, and speed over time horizons up to 20 s.

We used a single trajectory because additional trajectories did not significantly impact the overall RMSE value.

Emergent behavior was captured by extracting emergent features such as offroad duration, collision rate and hard brake rate. We calculate these rates by finding the amount of times a particular constraint is satisfied, and dividing by the total number of instances. For offroad duration, our constraint is when the vehicle is more than 1 m off one edge of the road. Collision rate is when there is a collision, which is easily accessible as one of our features. The threshold for a hard brake event is $-3\,\text{m}\,\text{s}^{-1}$.

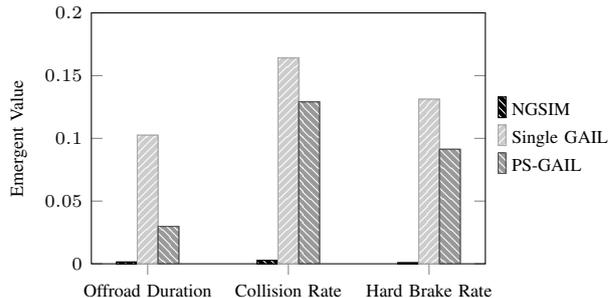

Fig. 2: Emergent values for each model.

### B. Results and Discussion

Fig. 1 shows root mean square error results for prediction horizons up to 20 s. These plots indicate that PS-GAIL captures expert behavior more faithfully than single-agent GAIL. This performance discrepancy is especially pronounced for longer prediction horizons, where the errors for single-agent policies begin to accumulate rapidly.

The superior performance of PS-GAIL is further illustrated by Fig. 2. These validation results empirically demonstrate that PS-GAIL policies are less likely to lead vehicles into collisions, extreme decelerations, and off-road driving. This serves as further illustration that the PS-GAIL training procedure encourages stabler interactions between agents, thereby making them less likely to encounter extreme or unlikely driving situations.

The difficulty of the multi-agent task scales with the number of agents controlled in the environment. At what number of controlled agents does the single-agent policy deteriorate beyond a usable point? We address this question in our third set of results. Fig. 3 shows the performance of the two models as a function of the number of agents controlled in the environment. For each number of agents, that many agents are replaced in the environment with the policy, while the remaining agents are left as originally recorded in NGSIM. The results indicate that while the single-agent policy deteriorates rapidly, the multi-agent policy declines in performance much more gradually.

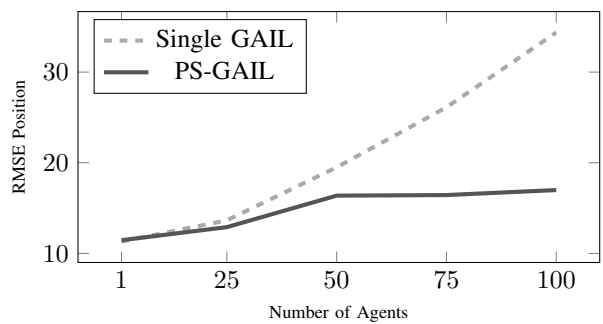

Fig. 3: This plot shows the average RMSE value across all timesteps of an episode as a function of the number of controlled agents. As the policy controls more vehicles, single-agent GAIL performance deteriorates rapidly, while PS-GAIL worsens at a much lower rate.

## VI. CONCLUSIONS

Validating the safety of autonomous vehicles in the real world is costly, dangerous, and time consuming. Performing this validation in simulated environments would help address these problems, but requires a highly realistic simulator. This research focused on building human driver models capable of interacting with other controlled vehicles in a manner representative of real data.

We proposed to train these models in a multi-agent setting using a parameter sharing approach that addresses scaling issues associated with multi-agent learning. Furthermore, we addressed challenges of instability in the learning environment through the use of a training curriculum. Experiments comparing the performance of this multi-agent model with existing single-agent models indicated that the former exhibits significantly more realistic behavior, particularly over longer time horizons.

Future work will investigate methods for improving model performance, and applying learned driver models. Three potential methods for improving model performance are (i) reward augmentation, (ii) applying learning algorithms that encourage more diverse behavior [13], and (iii) using a recurrent critic in order to account for partial observability. Ultimately, the goal in learning human driver models is to validate autonomous vehicles in simulation, and we hope to apply these models to that end in the future.

The code associated with this project and relevant videos can be found at https://github.com/sisl/ngsim_env.

## ACKNOWLEDGMENTS

Toyota Research Institute (TRI) provided funds to assist the authors with their research, but this article solely reflects the opinions and conclusions of its authors and not TRI or any other Toyota entity. This work was also supported in part by the Ford Motor Company and The Allstate Corporation. This work was partially performed by Blake Wulfe while at Adobe Systems Incorporated. We thank Jayesh Gupta for useful discussions.